\documentclass[11pt,a4paper]{article}
\usepackage[hyperref]{acl2021}
\usepackage{times}
\usepackage{latexsym}

\usepackage{footmisc}
 
\usepackage{microtype}
\usepackage{todonotes}
\presetkeys{todonotes}{inline}{}
\usepackage{booktabs}

\usepackage{scalerel}
\usepackage{amsmath}
\DeclareMathOperator*{\argmax}{arg\,max}

\aclfinalcopy 

\newcommand{\respace}{\vspace*{-.1cm}}
\title{Pivot Through English: Reliably Answering Multilingual Questions without Document Retrieval}

\author{Ivan Montero \\
  University of Washington \\
  \texttt{ivamon@cs.washington.edu} \\\And
  Shayne Longpre \\
  Apple Inc. \\
  \texttt{slongpre@apple.com} \\
  \AND
  Ni Lao \\
  Apple Inc. \\
  \texttt{ni\_lao@apple.com} \\
  \And
  Andrew J. Frank \\
  Apple Inc. \\
  \texttt{a\_frank@apple.com} \\
  \And
  Christopher DuBois \\
  Apple Inc. \\
  \texttt{cdubois@apple.com}\\
  }

\date{}

\begin{document}
\maketitle

\begin{abstract}
Existing methods for open-retrieval question answering in lower resource languages (LRLs) lag significantly behind English. 
They not only suffer from the shortcomings of non-English document retrieval, but are reliant on language-specific supervision for either the task or translation.
We formulate a task setup more realistic to available resources, that circumvents document retrieval to reliably transfer knowledge from English to lower resource languages.
Assuming a strong English question answering model or database, we compare and analyze methods that pivot through English: to map foreign queries to English and then English answers back to target language answers. 
Within this task setup we propose Reranked Multilingual Maximal Inner Product Search (RM-MIPS), akin to semantic similarity retrieval over the English training set with reranking, which outperforms the strongest baselines by 2.7\% on XQuAD and 6.2\% on MKQA.
Analysis demonstrates the particular efficacy of this strategy over state-of-the-art alternatives in challenging settings: low-resource languages, with extensive distractor data and query distribution misalignment.
Circumventing retrieval, our analysis shows this approach offers rapid answer generation to almost any language off-the-shelf, without the need for any additional training data in the target language.
\end{abstract}
\respace
\section{Introduction}
\respace
\label{sec:introduction}
Open-Retrieval question answering (ORQA) has seen extensive progress in English, significantly outperforming systems in lower resource languages (LRLs). 
This advantage is largely driven by the scale of labelled data and open source retrieval tools that exist predominantly for higher resource languages (HRLs) --- usually English. 

\begin{figure}[t]
\centering
\hspace{-2mm}\includegraphics[width=0.4625\textwidth]{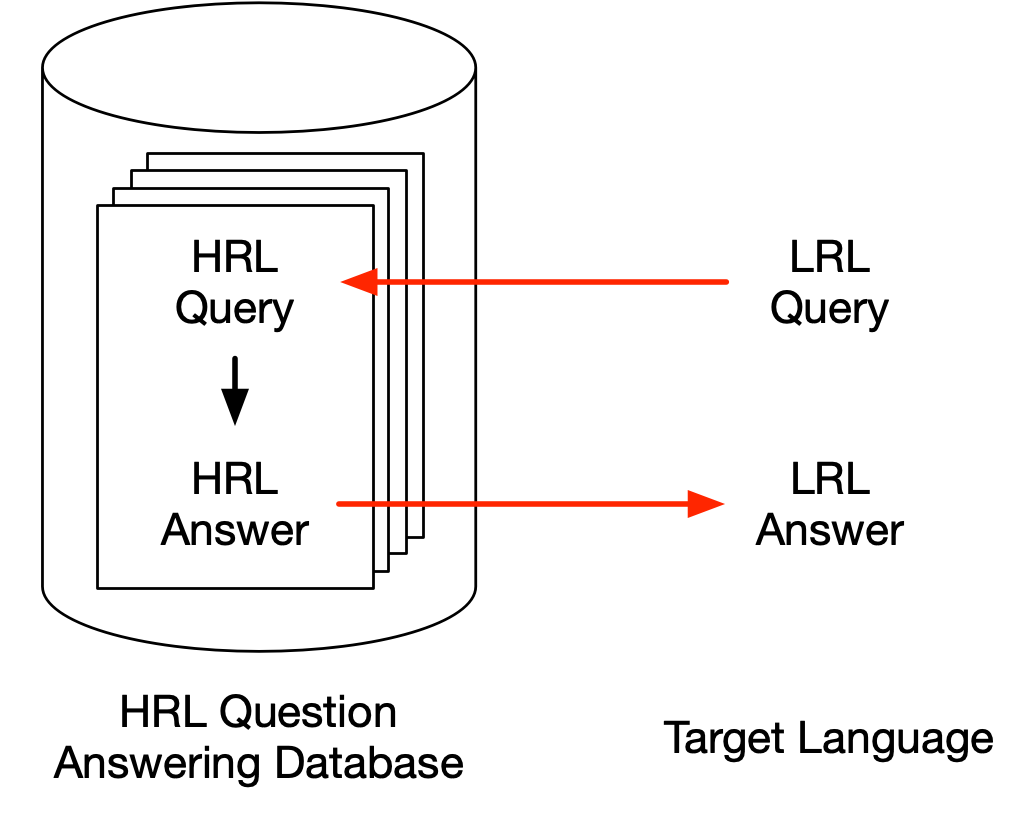}
\caption{\label{fig:overview} \textbf{Cross-Lingual Pivots (XLP):} We introduce the ``Cross Lingual Pivots" task, formulated as a solution to multilingual question answering that circumvents document retrieval in low resource languages (LRL). 
To answer LRL queries, approaches may leverage a question-answer system or database in a high resource language (HRL), such as English.}
\end{figure}

To remedy this discrepancy, recent work leverages English supervision to improve multilingual systems, either by simple translation or zero shot transfer \citep{asai2018multilingual, cui2019cross, charlet-etal-2020-cross}. 
While these approaches have helped generalize reading comprehension models to new languages, they are of limited practical use without reliable information retrieval in the target language, which they often implicitly assume.

\begin{figure*}[ht]
\centering
\hspace{-2mm}\includegraphics[width=\textwidth]{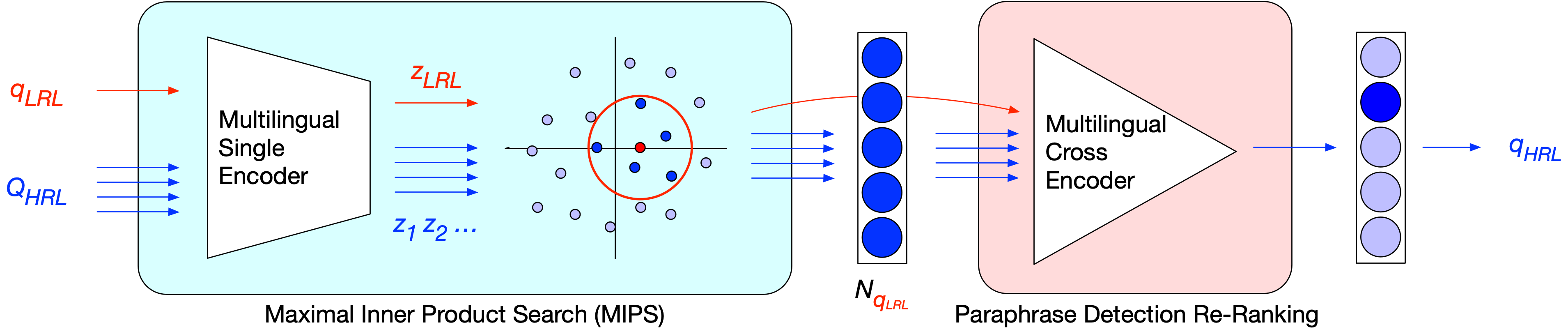}
\caption{\label{fig:xxen_diagram} \textbf{Reranked Multilingual Maximal Inner Product Search (RM-MIPS):} For the Cross-Lingual Pivots task, we propose an approach that maps the LRL query to a semantically equivalent HRL query, finds the appropriate HRL answer, then uses knowledge graph or machine translation to map the answer back to the target LRL.
Specifically, the first stage (in blue) uses multilingual single encoders for fast maximal inner product search (MIPS), and the second stage (in red) reranks the top k candidates using a more expressive multilingual cross-encoder that takes in the concatenation of the LRL query and candidate HRL query.}
\end{figure*}

In practice, we believe this assumption can be challenging to meet.
A new document index can be expensive to collect and maintain, and an effective retrieval stack typically requires language-specific labelled data, tokenization tools, manual heuristics, and curated domain blocklists \citep{fluhr1999multilingual, swapnil2014, manpreet2018}.
Consequently, we discard the common assumption of robust non-English document retrieval, for a more realistic one: that there exists a high-quality English database of query-answer string pairs.
We introduce and motivate the Cross-Lingual Pivots (\textbf{XLP}) task (Section~\ref{sec:task}), which we contend will accelerate progress in LRL question answering by reflecting these practical considerations.
This pivot task is similar to ``translate test" and ``MT-in-the-middle" paradigms \citep{10.3115/974147.974149, zitouni-florian-2008-mention, schneider-etal-2013-supersense} except for the availability of the high-resource language database, which allows for more sophisticated pivot approaches.
Figure~\ref{fig:overview} illustrates a generalized version of an XLP, where LRL queries may seek knowledge from any HRL with its own database.

For this task we combine and compare state-of-the-art methods in machine translation (``translate test") and cross-lingual semantic similarity, in order to map LRL queries to English, and then English answers back to the LRL target language.
In particular we examine how these methods are affected by certain factors: (a) whether the language is high, medium or low resource, (b) the magnitude of data in the HRL database, and (c) the degree of query distribution alignment between languages (i.e., the number of LRL queries that have matches in the HRL database). 

Lastly we propose a new approach to this task, motivated by recent dense nearest neighbour (kNN) models in English which achieve strong results in QA by simply searching for similar questions in the training set (or database in our case) \citep{lewis2020question}.
We leverage nearest neighbor semantic similarity search followed by cross-encoder reranking (see Figure~\ref{fig:xxen_diagram}), and refer to the technique as Reranked Multilingual Maximal Inner Product Search (\textbf{RM-MIPS}). 
Not only does this approach significantly improve upon ``Translate Test" (the most common pivot technique) and state-of-the-art paraphrase detection baselines, our analysis demonstrates it is more robust to lower resource languages, query distribution misalignment, and the size of the English database.

By circumventing document retrieval and task-specific supervision signals, this straightforward approach offers reliable answer generation to any language (present in pretraining) off-the-shelf. 
Furthermore, it can be re-purposed to obtain reliable training data in the target language, with fewer annotation artifacts, and is complementary to a standard end-to-end question answering system. 
We hope this analysis complements existing multilingual approaches, and facilitates adoption of more practical (but effective) methods to improve knowledge transfer from English into other languages.

We summarize our contributions as:
\begin{itemize}
  \item \textsc{XLP}: A more realistic task setup for practically expanding Multilingual OR-QA to lower resource languages.
  \item Comprehensive analysis of factors affecting XLP: (I) types of approaches (translation, paraphrasing) (II) language types, (III) database characteristics, and (IV) query distribution alignment.
  \item \textsc{RM-MIPS}: A flexible approach to XLP that beats strong (or state-of-the-art) baselines.
\end{itemize}

\respace
\section{Task: Cross-Lingual Pivots}
\respace
\label{sec:task}

The Open-Retrieval Question Answering (ORQA) task evaluates models' ability to answer information-seeking questions.
In a multilingual setting, the task is to produce answers in the same language as the query.
In some cases, queries may only find answers, or sufficient evidence, in a different language, due to \textit{informational asymmetries} \citep{miniwatts2011, https://doi.org/10.1002/asi.21577}.
To address this, \citet{asai2020xor} propose Cross-Lingual Open-Retrieval Question Answering (XORQA), similar to the Cross-Lingual Information Retrieval (CLIR) task, where a model needs to leverage intermediary information found in other languages, in order to serve an answer in the target language.
In practice, this intermediary language tends to be English, with the most ample resources and training data.

Building on these tasks, we believe there are other benefits to pivoting through high resource languages that have so far been overlooked, and consequently limited research that could more rapidly improve non-English QA.
These two benefits are (I) large query-answer databases have already been collected in English, both in academia \citep{joshi2017triviaqa} and in industry \citep{kwiatkowski2019natural}, and (II) it is often very expensive and challenging to replicate robust retrieval and passage reranking stacks in new languages \citep{fluhr1999multilingual, swapnil2014, manpreet2018}.~\footnote{While it is straightforward to adapt question answering ``reader" modules with zero-shot learning \citep{charlet-etal-2020-cross}, retrieval can be quite challenging. Not only is the underlying document index costly to expand and maintain for a new language \citep{swapnil2014}, but supervision signals collected in the target language are particularly important for dense retrieval and reranking systems which both serve as bottlenecks to downstream multilingual QA \citep{karpukhin2020dense}.
Additionally, real-world QA agents typically require human curated, language-specific infrastructure for retrieval, such as regular expressions, custom tokenization rules, and curated website blocklists.}
As a result, the English capabilities of question answering systems typically exceed those for non-English languages by large margins \citep{lewis2019mlqa, longpre2020mkqa, clark2020tydi}.

We would note that prior work suggests even without access to an English query-answer database, translation methods with an English document index and retrieval outperforms LRL retrieval for open-retrieval QA (see the end-to-end \textsc{XOR-Full} results in \citet{asai2020xor}).
This demonstrates the persistent weakness of non-English retrieval, and motivates alternatives approaches such as cross-lingual pivots.

To remedy this disparity, we believe attending to these two considerations would yield a more realistic task setup.
We propose the Cross-Lingual Pivots (XLPs) task, a variant of XORQA which assumes access to a query-answer database in English, as shown in Figure~\ref{fig:overview}.
Like multilingual ORQA, or XORQA, the task is to produce an answer $\hat{a}_{LRL}$ in the same ``Target" language as question $q_{LRL}$, evaluated by Exact Match of F1 token-overlap with the real answer $a_{LRL}$.
Instead of assuming access to a LRL document index or retrieval system (usually provided by the datasets), we assume access to an English database $D_{HRL}$ which simply maps English queries to their English answer text.
Leveraging this database, and circumventing LRL retrieval, we believe progress in this task will greatly accelerate multilingual capabilities of real question answering assistants.

\respace
\section{Re-Ranked Multilingual Maximal Inner Product Search}
\respace
\label{sec:methods}

For the first stage of the XLP task, our goal is to find an equivalent English query for a LRL query: ``Query Matching".
Competing approaches include Single Encoders and Cross Encoders, described further in section \ref{sec:query-matching}.
Single Encoders embed queries independently into a latent vector space, meaning each query $q_{EN}$ from the English database $Q_{EN}$ can be pre-embedded offline.
At inference time, the low resource query $q_{LRL}$ is embedded, then maximal inner product search (MIPS) finds the approximate closest query $q_{EN}$ among all $Q_{EN}$ by cosine similarity.
By comparison, Cross Encoders leverage cross-attention between $q_{LRL}$ and candidate match $q_{EN}$ at inference time, thus requiring $O(|Q_{EN}|)$ forward passes at inference time to find the best paraphrase.
While usually more accurate this is computationally infeasible for a large set of candidates.

We propose a method that combines both Single Encoders and Cross Encoders, which we refer to as Reranked Mulilingual Maximal Inner Product Search (RM-MIPS). 
The process, shown in Figure~\ref{fig:xxen_diagram}, first uses a multilingual sentence embedder with MIPS to isolate the top-k candidate similar queries, then uses the cross encoder to rerank the candidate paraphrases.
This approach reflects the Retrieve and Read paradigm common in OR-QA, but applies it to a multilingual setting for semantic similarity search.

The model first queries the English database using the Multilingual Single Encoder $\mathrm{SE}(q_i) = z_{i}$ to obtain the $k$-nearest English query neighbors $\mathcal{N}_{q_{LRL}} \subseteq Q_{EN}$ to the given query $q_{LRL}$ by cosine similarity.
\respace
$$\mathcal{N}_{q_{\scaleto{LRL}{3pt}}} = \argmax_{\{q_1, ..., q_k\} \subseteq Q_{\scaleto{EN}{4pt}}} \sum_{i=1}^{k} \mathrm{sim}(z_{\scaleto{LRL}{4pt}}, z_i)$$
\respace
Then, it uses the Multilingual Cross Encoder $\mathrm{CE}(q_1, q_2)$ to score the remaining set of queries $\mathcal{N}_{q_{\scaleto{LRL}{3pt}}}$ to obtain the final prediction. 

\respace
$$\mathrm{\textbf{RM-MIPS}}(q_{\scaleto{LRL}{4pt}}) = \argmax_{q_{\scaleto{EN}{4pt}} \in \mathcal{N}_{q_{\scaleto{LRL}{3pt}}}} \mathrm{CE}(q_{\scaleto{EN}{4pt}}, q_{\scaleto{LRL}{4pt}})$$
\respace

RM-MIPS($q_{\scaleto{LRL}{4pt}}$) proposes an equivalent English query $q_{\scaleto{EN}{4pt}}$, whose English answer can be pulled directly from the database.

\begin{table}[!htp] 
\centering
\begin{scriptsize}
\begin{tabular}{l|c|c}
\toprule
 & XQuAD & MKQA \\
\midrule
High & es, de, ru, zh & de, es, fr, it, ja, pl, pt, ru, zh\_cn \\
Medium & ar, tr, vi & ar, da, fi, he, hu, ko, nl, no, sv, tr, vi \\
Low & el, hi, th & km, ms, th, zh\_hk, zh\_tw  \\
\bottomrule
\end{tabular}
\end{scriptsize}
\caption{\label{tab:lang_groups}We evaluate cross-lingual pivot methods by language groups, divided into high, medium, and low resource according to Wikipedia coverage \citet{wu-dredze-2020-languages}. Note that due to greater language diversity, MKQA contains lower resource languages than XQuAD.}
\end{table}
\respace

\respace
\section{Experiments}
\respace
\label{sec:experiments}

We compare systems that leverage an English QA database to answer questions in lower resource languages.
Figure~\ref{fig:overview} illustrates a cross-lingual pivot (XLP), where the task is to map an incoming query from a low resource language to a query in the high resource language database (LRL $\rightarrow$ HRL, discussed in \ref{sec:query-matching}), and then a high resource language answer to a low resource language answer (HRL $\rightarrow$ LRL, discussed in \ref{sec:answer-gen}).

\begin{table*}[htb]
\centering
\begin{scriptsize}
\begin{tabular}{lllll|llll}
\toprule
\cmidrule{2-9}{MKQA + Natural Questions}
 & \multicolumn{4}{|c|}{LRL $\rightarrow$ HRL (Acc.)} & \multicolumn{4}{|c}{LRL $\rightarrow$ HRL $\rightarrow$ LRL (F1)} \\
\cmidrule{2-9}{Language Groups}
 & \multicolumn{1}{|c|}{All} & \multicolumn{1}{c}{High} & \multicolumn{1}{c}{Medium} & \multicolumn{1}{c}{Low} & \multicolumn{1}{|c|}{All} & \multicolumn{1}{c}{High} & \multicolumn{1}{c}{Medium} & \multicolumn{1}{c}{Low} \\
\midrule
NMT + MIPS & \multicolumn{1}{|l|}{74.4 $\pm$ 15.8} & 78.8 $\pm$ 13.3 & 78.3 $\pm$ 10.0 & 57.7 $\pm$ 19.0 & \multicolumn{1}{|l|}{65.8 $\pm$ 16.3} & 70.7 $\pm$ 14.5 & 69.9 $\pm$ 11.0 & 47.8 $\pm$ 17.0 \\
mUSE & \multicolumn{1}{|l|}{71.8 $\pm$ 21.2} & \textbf{88.2} $\pm$ 4.4 & 57.8 $\pm$ 20.4 & 73.2 $\pm$ 19.6 & \multicolumn{1}{|l|}{62.8 $\pm$ 18.3} & \textbf{77.8} $\pm$ 8.9 & 52.6 $\pm$ 16.9 & 58.2 $\pm$ 15.8 \\
LASER & \multicolumn{1}{|l|}{74.2 $\pm$ 15.0} & 70.0 $\pm$ 14.6 & 82.6 $\pm$ 8.5 & 63.3 $\pm$ 16.8 & \multicolumn{1}{|l|}{65.4 $\pm$ 15.4} & 62.8 $\pm$ 14.3 & 73.6 $\pm$ 9.4 & 52.0 $\pm$ 16.6 \\
Single Encoder (XLM-R) & \multicolumn{1}{|l|}{73.0 $\pm$ 6.8} & 72.6 $\pm$ 3.7 & 73.4 $\pm$ 8.3 & 72.6 $\pm$ 7.3 & \multicolumn{1}{|l|}{63.2 $\pm$ 8.1} & 63.9 $\pm$ 4.9 & 65.4 $\pm$ 8.9 & 57.1 $\pm$ 8.0 \\
\midrule
RM-MIPS (mUSE) & \multicolumn{1}{|l|}{78.2 $\pm$ 12.5} & 86.9 $\pm$ 3.1 & 71.9 $\pm$ 12.5 & 76.7 $\pm$ 14.0 & \multicolumn{1}{|l|}{68.1 $\pm$ 12.4} & 76.3 $\pm$ 8.0 & 64.9 $\pm$ 11.3 & 60.4 $\pm$ 12.7 \\
RM-MIPS (LASER) & \multicolumn{1}{|l|}{80.1 $\pm$ 9.4} & 79.5 $\pm$ 7.8 & \textbf{83.7} $\pm$ 5.6 & 73.1 $\pm$ 13.6 & \multicolumn{1}{|l|}{69.4 $\pm$ 11.2} & 70.0 $\pm$ 9.3 & 74.1 $\pm$ 7.3 & 57.8 $\pm$ 13.2 \\
RM-MIPS (XLM-R) & \multicolumn{1}{|l|}{\textbf{83.5} $\pm$ 5.2} & 84.9 $\pm$ 2.7 & \textbf{83.7} $\pm$ 5.7 & \textbf{80.7} $\pm$ 6.1 & \multicolumn{1}{|l|}{\textbf{72.0} $\pm$ 9.3} & 74.7 $\pm$ 7.6 & \textbf{74.2} $\pm$ 7.7 & \textbf{62.7} $\pm$ 9.5 \\

\midrule
\textit{Perfect LRL $\rightarrow$ HRL} & \multicolumn{1}{|l|}{ - } & - & - & - & \multicolumn{1}{|l|}{90.1 $\pm$ 7.3} & 91.8 $\pm$ 7.1 & 92.4 $\pm$ 4.2 & 81.9 $\pm$ 7.5 \\
\bottomrule

\end{tabular}
\end{scriptsize}
\caption{\label{tab:mkqa}
\textbf{MKQA results by language group with MKQA + Natural Questions as the HRL Database}: 
(left) the accuracy for the LRL $\rightarrow$ HRL Query Matching stage;
(right) the F1 scores for the End-to-End XLP task, using WikiData translation for Answer Translation; and 
(bottom) the F1 score only for Wikidata translation, assuming Query Matching (LRL $\rightarrow$ HRL) was perfect. 
Macro standard deviation are computed for language groups ($\pm$).
The difference between all method pairs are significant.
}
\end{table*}

\respace

\begin{table*}[htb]
\centering
\begin{scriptsize}
\begin{tabular}{lllll|llll}
\toprule
\cmidrule{2-9}{XQuAD + SQuAD}
 & \multicolumn{4}{|c|}{LRL $\rightarrow$ HRL (Acc.)} & \multicolumn{4}{|c}{LRL $\rightarrow$ HRL $\rightarrow$ LRL (F1)} \\
\cmidrule{2-9}{Language Group}
 & \multicolumn{1}{|c|}{All} & \multicolumn{1}{c}{High} & \multicolumn{1}{c}{Medium} & \multicolumn{1}{c}{Low} & \multicolumn{1}{|c|}{All} & \multicolumn{1}{c}{High} & \multicolumn{1}{c}{Medium} & \multicolumn{1}{c}{Low} \\
\midrule

NMT + MIPS & \multicolumn{1}{|l|}{77.7 $\pm$ 14.4} & 78.4 $\pm$ 21.4 & 76.5 $\pm$ 4.7 & 78.0 $\pm$ 8.0 & \multicolumn{1}{|l|}{24.5 $\pm$ 12.0} & 28.8 $\pm$ 17.3 & 24.5 $\pm$ 3.3 & 18.7 $\pm$ 3.8 \\
mUSE & \multicolumn{1}{|l|}{68.0 $\pm$ 38.5} & \textbf{94.5} $\pm$ 3.0 & 66.4 $\pm$ 34.5 & 34.2 $\pm$ 40.7 & \multicolumn{1}{|l|}{21.1 $\pm$ 15.8} & \textbf{31.9} $\pm$ 15.6 & 20.3 $\pm$ 9.8 & 7.3 $\pm$ 7.8 \\
LASER & \multicolumn{1}{|l|}{46.7 $\pm$ 24.9} & 54.7 $\pm$ 24.3 & 63.9 $\pm$ 1.6 & 18.8 $\pm$ 10.9 & \multicolumn{1}{|l|}{15.2 $\pm$ 11.6} & 20.1 $\pm$ 14.1 & 19.9 $\pm$ 2.3 & 4.1 $\pm$ 2.3 \\
Single Encoder (XLM-R) & \multicolumn{1}{|l|}{81.4 $\pm$ 6.2} & 85.1 $\pm$ 1.9 & 79.4 $\pm$ 9.4 & 78.6 $\pm$ 2.2 & \multicolumn{1}{|l|}{24.3 $\pm$ 10.8} & 29.1 $\pm$ 14.4 & 24.5 $\pm$ 5.3 & 17.7 $\pm$ 3.0 \\
\midrule
RM-MIPS (mUSE) & \multicolumn{1}{|l|}{72.0 $\pm$ 34.0} & \textbf{94.4} $\pm$ 2.5 & 75.1 $\pm$ 25.4 & 39.1 $\pm$ 37.8 & \multicolumn{1}{|l|}{22.4 $\pm$ 14.7} & \textbf{31.8} $\pm$ 15.4 & 23.7 $\pm$ 6.0 & 8.5 $\pm$ 6.9 \\
RM-MIPS (LASER) & \multicolumn{1}{|l|}{69.2 $\pm$ 23.7} & 77.5 $\pm$ 14.8 & 85.4 $\pm$ 3.0 & 41.9 $\pm$ 21.8 & \multicolumn{1}{|l|}{21.2 $\pm$ 12.3} & 26.7 $\pm$ 14.3 & 26.0 $\pm$ 3.1 & 9.2 $\pm$ 4.0 \\
RM-MIPS (XLM-R) & \multicolumn{1}{|l|}{\textbf{92.2} $\pm$ 2.4} & 93.4 $\pm$ 1.7 & \textbf{90.4} $\pm$ 2.7 & \textbf{92.3} $\pm$ 1.4 & \multicolumn{1}{|l|}{\textbf{27.2} $\pm$ 10.8} & 31.5 $\pm$ 15.2 & \textbf{27.4} $\pm$ 3.1 & \textbf{21.2} $\pm$ 2.8 \\

\midrule
\textit{Perfect LRL $\rightarrow$ HRL} & \multicolumn{1}{|l|}{ - } & - & - & - & \multicolumn{1}{|l|}{46.6 $\pm$ 13.1} &  51.0 $\pm$ 15.5 & 51.2 $\pm$ 5.0 & 36.3 $\pm$ 8.4 \\
\bottomrule

\end{tabular}
\end{scriptsize}
\caption{\label{tab:xquad}
\textbf{XQuAD results by language group with XQuAD + SQuAD as the HRL Database}: 
(left) the accuracy for the LRL $\rightarrow$ HRL Query Matching stage;
(right) the F1 scores for the End-to-End XLP task, using machine translation to translate answers from HRL $\rightarrow$ LRL; and 
(bottom) the F1 score only for Wikidata translation, assuming Query Matching (LRL $\rightarrow$ HRL) was perfect. 
Macro standard deviation are computed for language groups ($\pm$).
The difference between all method pairs are significant.
}
\end{table*}

\respace

\subsection{Datasets}
\respace
We provide an overview of the question answering and paraphrase datasets relevant to our study.

\subsubsection{Question Answering}
\respace

To assess cross-lingual pivots, we consider multilingual OR-QA evaluation sets that (a) contain a diverse set of language families, and (b) have ``parallel" questions across all of these languages.
The latter property affords us the opportunity to change the distributional overlap and analyze its effect (\ref{sec:answer-dropout}).

\respace
\paragraph{XQuAD}
\citet{artetxe2019cross} human translate ~1.2k SQuAD examples \citep{rajpurkar2016squad} into 10 other languages. We use all of SQuAD (100k+) as the associated English database, such that only ~1\% of database queries are represented in the LRL evaluation set.

\respace
\paragraph{MKQA} \citet{longpre2020mkqa} human translate 10k examples from the Natural Questions \citep{kwiatkowski2019natural} dataset to 25 other languages. We use the rest of the Open Natural Questions training set (84k) as the associated English  database, such that only ~10.6\% of the database queries are represented in the LRL evaluation set\footnote{Open Natural Questions train set found here: \url{https://github.com/google-research-datasets/natural-questions/tree/master/nq_open}}.

\subsubsection{Paraphrase Detection}
\respace
\label{sec:paraphrase-datasets}

To detect paraphrases between LRL queries and HRL queries we train multilingual sentence embedding models with a mix of the following paraphrase datasets.

\respace
\paragraph{PAWS-X}
\citet{yang2019paws} machine translate ~49k examples from the PAWS \citep{zhang2019paws} dataset to six other languages. 
This dataset provides both positive and negative paraphrase examples.

\respace
\paragraph{Quora Question Pairs} 
\citet{sharma2019natural} provide English question pair examples from Quora; we use the 384k examples from the training split of \citet{wang2017bilateral}.
This dataset provides both positive and negative examples of English paraphrases.

\subsection{Query Matching Baselines: LRL Query $\rightarrow$ HRL Query}
\respace
\label{sec:query-matching}

We consider a combination of translation techniques and cross-lingual sentence encoders to find semantically equivalent queries across languages.
We select from pretrained models which report strong results on similar multilingual tasks, or finetune representations for our task using publicly available paraphrase datasets (\ref{sec:paraphrase-datasets}). 
Each finetuned model receives basic hyperparameter tuning over the learning rate and the ratio of training data from PAWS-X and QQP.\footnote{We used an optimal learning rate of 1e-5, and training data ratio of 75\% PAWS-X and 25\% QQP.}

\respace
\paragraph{NMT + MIPS}
\label{sec:nmt-mips}
We use a many-to-many, Transformer-based \citep{vaswani2017attention}, encoder-decoder neural machine translation system, trained on the OPUS multilingual corpus covering 100 languages \citep{zhang2020improving}. 
To match the translation to an English query, we use the Universal Sentence Encoder (USE) \citep{cer2018universal} to perform maximal inner product search (MIPS).

\respace
\paragraph{Pretrained Single Encoders}
We consider pre-trained multilingual sentence encoders for sentence retrieval.
We explore mUSE\footnote{mUSE was only trained on the following 16 languages: ar, ch\_cn, ch\_tw, en, fr, de, it, ja, ko da, pl, pt, es, th, tr ru} \citep{yang2019multilingual}, LASER \citep{artetxe2019massively}, and m-SentenceBERT \cite{reimers2019sentence}.

\respace
\paragraph{Finetuned Single Encoders}
We finetune transformer encoders to embed sentences, per \citet{reimers2019sentence}.
We use the softmax loss over the combination of $[x; y; |x-y|]$ from \citet{conneau2017supervised} and mean pool over the final encoder representations to obtain the final sentence representation.
We use XLM-R Large as the base encoder \citep{conneau2019unsupervised}.

\respace
\paragraph{Cross Encoders}
We finetune XLM-R Large \citep{conneau2019unsupervised} which is pretrained using the multilingual masked language modelling (MLM) objective.\footnote{\label{huggingface}We use the pretrained Transformer encoder implementations in the Huggingface library \citep{Wolf2019HuggingFacesTS}.}
For classification, a pair of sentences are given as input for classification, taking advantage of cross-attention between sentences.

\subsection{Answer Translation: HRL Answer $\rightarrow$ LRL Answer}
\respace
\label{sec:answer-gen}

Once we've found an English (HRL) query using RM-MIPS, or one of our ``Query Matching" baselines, we can use the English database to lookup the English answer.
Our final step is to generate an equivalent answer in the target (LRL) language.

We explore straightforward methods of answer generation, including basic neural machine translation (NMT), and WikiData entity translation.

\respace
\paragraph{Machine Translation}
For NMT we use our many-to-many neural machine translation as described in Section~\ref{sec:nmt-mips}.

\begin{figure*}[htb]
\centerline{
\hspace{-2mm}\includegraphics[width=\textwidth]{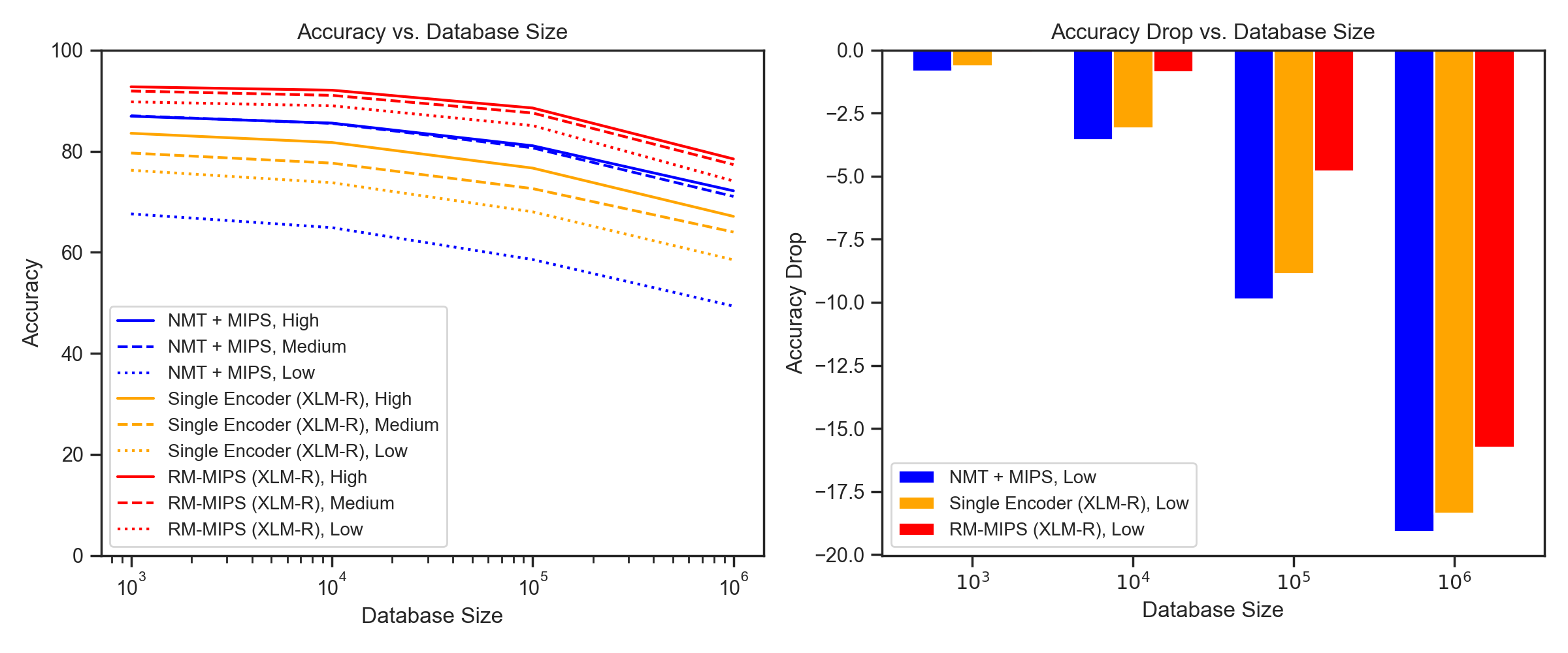}}
\caption{\label{fig:distractor} \textbf{Effect of Database Size on LRL $\rightarrow$ HRL.} 
Left: Query Matching accuracy of the strongest methods on different language groups as the amount of ``unaligned" queries in the English database increases. 
Right: The accuracy drop of the different methods on low resource languages as the amount of queries in the English database increases beyond the original parallel count.}
\end{figure*}


\respace

\paragraph{WikiData Entity Translation}
We propose our WikiData entity translation method for QA datasets with primarily entity type answers that would likely appear in the WikiData knowledge graph \citep{10.1145/2629489}.\footnote{\url{https://www.wikidata.org}}
This method uses a named entity recognizer (NER) with a WikiData entity linker to find an entity \citep{spacy2}. \footnote{\url{https://github.com/explosion/spaCy}}
We train our own entity linker on the public WikiData entity dump according to spaCy's instructions.
If a WikiData entity is found, its structured metadata often contains the equivalent term in the target language, localized to the relevant script/alphabet.
For our implementation, when a WikiData entity is not found, or its translation is not available in the target language, we simply return the English answer.

For XQuAD end-to-end experiments we find straightforward machine translation works best, whereas for MKQA, which contains more short, entity-type answers, we find WikiData Entity Translation works best.
We report results using these simple methods and leave more sophisticated combinations or improvements to future work.

\respace
\section{Results}
\respace
\label{sec:results}

\subsection{End-To-End Results}
\respace

We benchmark the performance of the cross-lingual pivot methods on XQuAD and MKQA.
To simulate a realistic setting, we add all the English questions from SQuAD to the English database used in the XQuAD experiments.
Similarly we add all of Natural Questions queries (not just those aligned across languages) in the MKQA experiments.
For each experiment we group the languages into high, medium, and low resource, as shown in Table~\ref{tab:lang_groups}, according to \citet{wu-dredze-2020-languages}.
Tables~\ref{tab:mkqa} and ~\ref{tab:xquad} present the mean performance by language group, for query matching (LRL $\rightarrow$ HRL), and end-to-end results (LRL $\rightarrow$ HRL $\rightarrow$ LRL), query matching and answer translation ins sequence.

Among the models, RM-MIPS typically outperform baselines, particularly on lower resource languages.
We find the reranking component in particular offers significant improvements over the non-reranked sentence encoding approaches in low resource settings, where we believe sentence embeddings are most inconsistent in their performance.
For instance, RM-MIPS (LASER) outperforms LASER by $5.7\%$ on the Lowest resource E2E MKQA task, and $4.0\%$ across all languages.
The margins are even larger between RM-MIPS (mUSE) and mUSE as well as RM-MIPS (XLM-R) and XLM-R).

\begin{figure*}[htb]
    \centering
    \includegraphics[width=\textwidth]{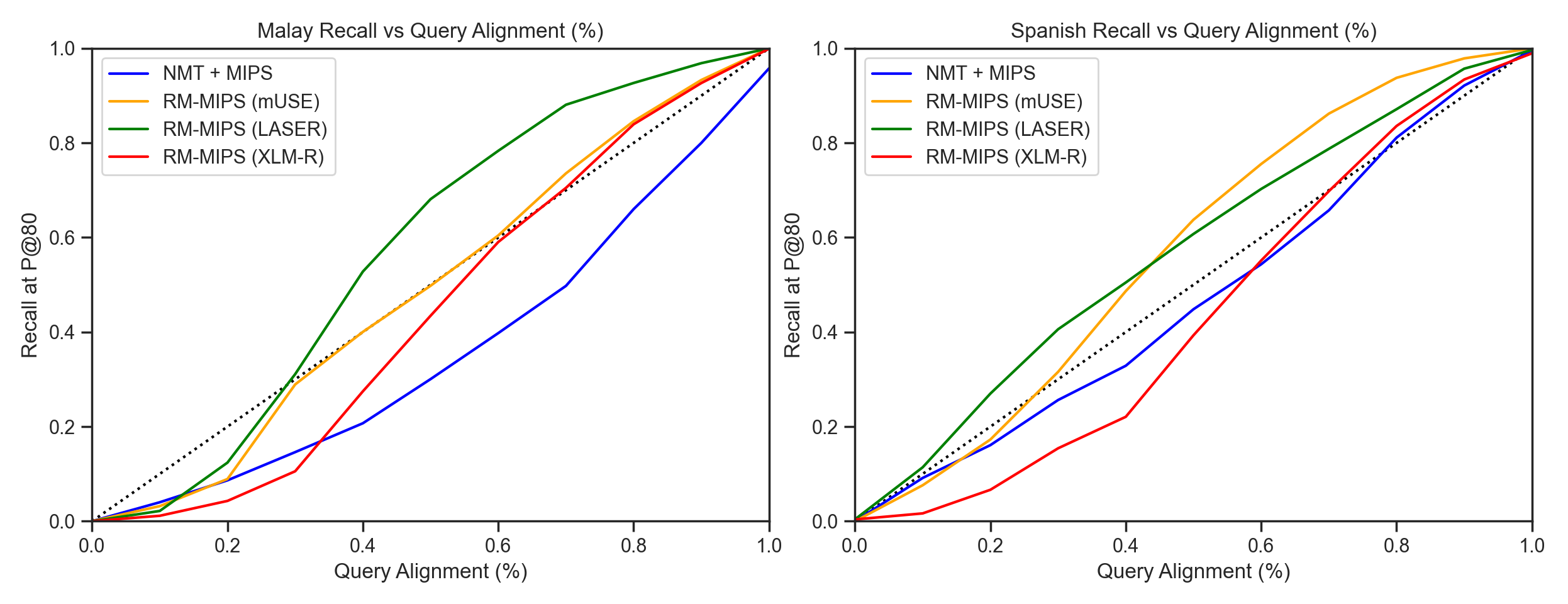}
    \caption{\label{fig:dropout} \textbf{Effects of Query Alignment on MKQA end-to-end Performance:} At a target precision of 80\%, the end-to-end Malay (left) and Spanish (right) recall are plotted for each degree of query alignment.
    The query alignment axis indicates the percentage of 10k queries with parallel matches retained in the English database.}
\end{figure*}
\respace

For certain high resource languages, mUSE performs particularly strongly, and for XQuAD languages, LASER performs poorly.
Accordingly, the choice of sentence encoder (and it's language proportions in pretraining) is important in optimizing for the cross-lingual pivot task.
The modularity of RM-MIPS offers this flexibility, as the first stage multiligual encoder can be swapped out: we present results for LASER, mUSE, and XLM-R.

Comparing query matching accuracy (left) and end-to-end F1 (right) in Tables~\ref{tab:mkqa} and ~\ref{tab:xquad} measures the performance drop due to answer translation (HRL $\rightarrow$ LRL, see section~\ref{sec:answer-gen} for details).
We see this drop is quite small for MKQA as compared to XQuAD.
Similarly, the ``Perfect LRL $\rightarrow$ HRL" measures the Answer Translation stage on all queries, showing XQuAD's machine translation for answers is much lower than MKQA's Wikidata translation for answers.
This observation indicates that (a) Wikidata translation is particularly strong, and (b) cross-lingual pivot techniques are particularly useful for datasets with frequent entity, date, or numeric-style answers, that can be translated with Wikidata, as seen in MKQA.
Another potential factor in the performance difference between MKQA and XQuAD is that MKQA contains naturally occurring questions, whereas XQuAD does not.
Despite the lower mean end-to-end performance for XQuAD, this cross-lingual pivot can still be used alongside traditional methods, and can be calibrated for high precision/low coverage by abstaining from answering questions that are Wikidata translatable.

One other noteable advantage of paraphrase-based pivot approaches, is that no LRL-specific annotated training data is required.
A question answering system in the target language requires in-language annotated data, or an NMT system from English.
Traditional NMT ``translate test" or ``MT-in-the-middle" \citep{asai2018multilingual, 10.3115/974147.974149, schneider-etal-2013-supersense} approaches also require annotated parallel data to train.
RM-MIPS and our other paraphrase baselines observe monolingual corpora at pre-training time, and only select language pairs during fine-tuning (those present in PAWS-X), and yet these models still perform well on XLP even for non-PAWS-X languages.

\subsection{Database Size}
\respace

To understand the impact of database size on the query matching process, we assemble a larger database with MSMARCO (800k), SQuAD (100k), and Open-NaturalQuestions (90k).
Note that none of the models are explicitly tuned to MKQA, and since MSMARCO and Open-NQ comprise natural user queries (from the same or similar distribution), we believe these are challenging ``distractors".
In Figure~\ref{fig:distractor} we plot accuracy of the most performant models from Tables~\ref{tab:mkqa} and ~\ref{tab:xquad} on each of the high, medium, and low resource language groups over different sizes of database on MKQA.
We report the initial stage query matching (LRL $\rightarrow$ HRL) to isolate individual model matching performance.
We observe that RM-MIPS degrades less quickly with database size than competing methods, and that it degrades less with the resourcefulness of the language group.

\subsection{Query Alignment}
\respace
\label{sec:answer-dropout}

In some cases, incoming LRL queries may not have a corresponding semantic match in the HRL database.
To assess the impact of this, we vary the percentage of queries that have a corresponding match by dropping out their parallel example in the English database (in increments of 10\%).
In Figures~\ref{fig:dropout} we report the median end-to-end recall scores over five different random seeds, at each level of query alignment (x-axis).
At each level of answer query alignment we recompute a No Answer confidence threshold for a target precision of 80\%.
Due to computational restraints, we select one low resource (Malay) and one high resource language (Spanish) to report results on.
We find that even calibrated for high precision (a target of 80\%) the cross-lingual pivot methods can maintain proportional, and occasionally higher, coverage to the degree of query misalignment.
RM-MIPS methods in particular can \textit{outperform} proportional coverage to alignment (the dotted black line on the diagonal) by sourcing answers from similar queries in the database to those dropped out.
Consequently, a practitioner can maintain high precision and respectable recall by selecting a threshold for any degree of query misalignment observed in their test distribution.

The primary limitation of RM-MIPS, or other pivot-oriented approaches, is that their performance is bounded by the degree of query alignment. 
However, QA systems still fail to replicate their English answer coverage in LRLs \citep{longpre2020mkqa}, and so we expect pivot techniques to remain essential until this gap narrows completely.

\respace
\section{Related Work}
\respace
\label{sec:related-work}

\paragraph{Cross-Lingual Modeling}
Multilingual BERT \citep{devlin2019bert}, XLM \citep{lample2019cross}, and XLM-R \cite{conneau2019unsupervised} use masked language modeling (MLM) to share embeddings across languages.
\citet{artetxe2019massively} introduce LASER, a language-agnostic sentence embedder trained using many-to-many machine translation.
\citet{yang2019multilingual} extend \citet{cer2018universal} in a multilingual setting by following \citet{chidambaram2019learning} to train a multi-task dual-encoder model (mUSE).
These multilingual encoders are often used for semantic similarity tasks.
\citet{reimers2019sentence} propose finetuning pooled BERT token representations (Sentence-BERT), and \citet{reimers2020making} extend with knowledge distillation to encourage vector similarity among translations.
Other methods improve multilingual transfer via language alignment \citep{roy2020lareqa, mulcaire2019polyglot, schuster2019cross} or combining machine translation with multilingual encoders \citep{fang2020filter, cui2019cross}.

\respace
\paragraph{Multilingual Question Answering}
Efforts to explore multilingual question answering include MLQA \citep{lewis2019mlqa}, XQuAD \citep{artetxe2019cross}, MKQA \citep{longpre2020mkqa}, TyDi \citep{clark2020tydi} and XORQA \citep{asai2020xor}. 

Prior work in multilingual QA achieves strong results combining neural machine translation and multilingual representations via \textbf{Translate-Test}, \textbf{Translate-Train}, or \textbf{Zero Shot} approaches \citep{asai2018multilingual, cui2019cross, charlet-etal-2020-cross}.
This work focuses on \textit{extracting} the answer from a multilingual passage \citep{cui2019cross, asai2018multilingual}, assuming passages are provided.
 
\respace
\paragraph{Improving Low Resource With High Resource} 
Efforts to improve performance on low-resource languages usually explore language alignment or transfer learning.
\citet{chung2017supervised} find supervised and unsupervised improvements in transfer learning when finetuning from a language specific model, and \citet{lee2019cross} leverage a GAN-inspired discriminator \citep{goodfellow2014generative} to enforce language-agnostic representations.
Aligning vector spaces of text representations in existing models \citep{conneau2017word, schuster2019cross, mikolov2013exploiting} remains a promising direction. 
Leveraging high resource data has also been studied in sequence labeling \citep{xie2018neural, plank2018distant, schuster2019cross} and machine translation \citep{johnson2017google,zhang2020improving}.

\respace
\paragraph{Paraphrase Detection} 
The paraphrase detection task determines whether two sentences are semantically equivalent. 
Popular paraphrase datasets include Quora Question Pairs \citep{sharma2019natural}, MRPC \citep{dolan2005automatically}, and STS-B \citep{Cer_2017}.
The adversarially constructed PAWS dataset \citet{zhang2019paws} was translated to 6 langauges, offering a multilingual option, PAWS-X \citet{yang2019paws}.
In a multilingual setting, an auxiliary paraphrase detection (or nearest neighbour) component, over a datastore of training examples, has been shown to greatly improve performance for neural machine translation \citep{khandelwal2020nearest}.

\respace
\section{Conclusion}
\respace
\label{sec:conclusion}
In conclusion, we formulate a task to cross-lingual open-retrieval question answering more realistic to the constraints and challenges faced by practitioners expanding their systems' capabilities beyond English.
Leveraging access to a large English training set our method of query retrieval followed by reranking greatly outperforms strong baseline methods.
Our analysis compares multiple methods of leveraging this English expertise and concludes our two-stage approach transfers better to lower resource languages, and is more robust in the presence of extensive distractor data and query distribution misalignment.
Circumventing retrieval, this approach offers fast online or offline answer generation to hundreds of languages straight off-the-shelf, without any additional training data for the target language.

We hope this analysis will promote creative methods in multilingual knowledge transfer, and the cross-lingual pivots task will encourage researchers to pursue problem formulations better informed by the needs of existing systems.
In particular, leveraging many location and culturally-specific query knowledge bases, with cross-lingual pivots across many languages is an exciting extension of this work.


\bibliography{anthology,acl2021}

\begin{thebibliography}{56}
\expandafter\ifx\csname natexlab\endcsname\relax\def\natexlab#1{#1}\fi

\bibitem[{Artetxe et~al.(2019)Artetxe, Ruder, and Yogatama}]{artetxe2019cross}
Mikel Artetxe, Sebastian Ruder, and Dani Yogatama. 2019.
\newblock On the cross-lingual transferability of monolingual representations.
\newblock \emph{arXiv preprint arXiv:1910.11856}.

\bibitem[{Artetxe and Schwenk(2019)}]{artetxe2019massively}
Mikel Artetxe and Holger Schwenk. 2019.
\newblock Massively multilingual sentence embeddings for zero-shot
  cross-lingual transfer and beyond.
\newblock \emph{Transactions of the Association for Computational Linguistics},
  7:597--610.

\bibitem[{Asai et~al.(2018)Asai, Eriguchi, Hashimoto, and
  Tsuruoka}]{asai2018multilingual}
Akari Asai, Akiko Eriguchi, Kazuma Hashimoto, and Yoshimasa Tsuruoka. 2018.
\newblock Multilingual extractive reading comprehension by runtime machine
  translation.
\newblock \emph{arXiv preprint arXiv:1809.03275}.

\bibitem[{Asai et~al.(2020)Asai, Kasai, Clark, Lee, Choi, and
  Hajishirzi}]{asai2020xor}
Akari Asai, Jungo Kasai, Jonathan~H Clark, Kenton Lee, Eunsol Choi, and
  Hannaneh Hajishirzi. 2020.
\newblock Xor qa: Cross-lingual open-retrieval question answering.
\newblock \emph{arXiv preprint arXiv:2010.11856}.

\bibitem[{Callahan and Herring(2011)}]{https://doi.org/10.1002/asi.21577}
Ewa~S. Callahan and Susan~C. Herring. 2011.
\newblock \href {https://doi.org/https://doi.org/10.1002/asi.21577} {Cultural
  bias in wikipedia content on famous persons}.
\newblock \emph{Journal of the American Society for Information Science and
  Technology}, 62(10):1899--1915.

\bibitem[{Cer et~al.(2017)Cer, Diab, Agirre, Lopez-Gazpio, and
  Specia}]{Cer_2017}
Daniel Cer, Mona Diab, Eneko Agirre, Inigo Lopez-Gazpio, and Lucia Specia.
  2017.
\newblock \href {https://doi.org/10.18653/v1/s17-2001} {Semeval-2017 task 1:
  Semantic textual similarity multilingual and crosslingual focused
  evaluation}.
\newblock \emph{Proceedings of the 11th International Workshop on Semantic
  Evaluation (SemEval-2017)}.

\bibitem[{Cer et~al.(2018)Cer, Yang, Kong, Hua, Limtiaco, John, Constant,
  Guajardo-Cespedes, Yuan, Tar et~al.}]{cer2018universal}
Daniel Cer, Yinfei Yang, Sheng-yi Kong, Nan Hua, Nicole Limtiaco, Rhomni~St
  John, Noah Constant, Mario Guajardo-Cespedes, Steve Yuan, Chris Tar, et~al.
  2018.
\newblock Universal sentence encoder for english.
\newblock In \emph{Proceedings of the 2018 Conference on Empirical Methods in
  Natural Language Processing: System Demonstrations}, pages 169--174.

\bibitem[{Charlet et~al.(2020)Charlet, Damnati, Bechet, Marzinotto, and
  Heinecke}]{charlet-etal-2020-cross}
Delphine Charlet, Geraldine Damnati, Frederic Bechet, Gabriel Marzinotto, and
  Johannes Heinecke. 2020.
\newblock \href {https://www.aclweb.org/anthology/2020.lrec-1.674}
  {Cross-lingual and cross-domain evaluation of machine reading comprehension
  with squad and {CALOR}-quest corpora}.
\newblock In \emph{Proceedings of the 12th Language Resources and Evaluation
  Conference}, pages 5491--5497, Marseille, France. European Language Resources
  Association.

\bibitem[{Chaudhari(2014)}]{swapnil2014}
Swapnil Chaudhari. 2014.
\newblock \href
  {http://www.cfilt.iitb.ac.in/resources/surveys/Swapnil-Cross-lingual-Information-Retrieval.pdf}
  {Cross lingual information retrieval}.
\newblock \emph{Center for Indian Language Technology}.

\bibitem[{Chidambaram et~al.(2019)Chidambaram, Yang, Cer, Yuan, Sung, Strope,
  and Kurzweil}]{chidambaram2019learning}
Muthu Chidambaram, Yinfei Yang, Daniel Cer, Steve Yuan, Yunhsuan Sung, Brian
  Strope, and Ray Kurzweil. 2019.
\newblock Learning cross-lingual sentence representations via a multi-task
  dual-encoder model.
\newblock In \emph{Proceedings of the 4th Workshop on Representation Learning
  for NLP (RepL4NLP-2019)}, pages 250--259.

\bibitem[{Chung et~al.(2017)Chung, Lee, and Glass}]{chung2017supervised}
Yu-An Chung, Hung-Yi Lee, and James Glass. 2017.
\newblock Supervised and unsupervised transfer learning for question answering.
\newblock \emph{arXiv preprint arXiv:1711.05345}.

\bibitem[{Clark et~al.(2020)Clark, Choi, Collins, Garrette, Kwiatkowski,
  Nikolaev, and Palomaki}]{clark2020tydi}
Jonathan~H Clark, Eunsol Choi, Michael Collins, Dan Garrette, Tom Kwiatkowski,
  Vitaly Nikolaev, and Jennimaria Palomaki. 2020.
\newblock Tydi qa: A benchmark for information-seeking question answering in
  typologically diverse languages.
\newblock \emph{arXiv preprint arXiv:2003.05002}.

\bibitem[{Conneau et~al.(2019)Conneau, Khandelwal, Goyal, Chaudhary, Wenzek,
  Guzm{\'a}n, Grave, Ott, Zettlemoyer, and Stoyanov}]{conneau2019unsupervised}
Alexis Conneau, Kartikay Khandelwal, Naman Goyal, Vishrav Chaudhary, Guillaume
  Wenzek, Francisco Guzm{\'a}n, Edouard Grave, Myle Ott, Luke Zettlemoyer, and
  Veselin Stoyanov. 2019.
\newblock Unsupervised cross-lingual representation learning at scale.
\newblock \emph{arXiv preprint arXiv:1911.02116}.

\bibitem[{Conneau et~al.(2017{\natexlab{a}})Conneau, Kiela, Schwenk, Barrault,
  and Bordes}]{conneau2017supervised}
Alexis Conneau, Douwe Kiela, Holger Schwenk, Lo{\"\i}c Barrault, and Antoine
  Bordes. 2017{\natexlab{a}}.
\newblock Supervised learning of universal sentence representations from
  natural language inference data.
\newblock In \emph{Proceedings of the 2017 Conference on Empirical Methods in
  Natural Language Processing}, pages 670--680.

\bibitem[{Conneau et~al.(2017{\natexlab{b}})Conneau, Lample, Ranzato, Denoyer,
  and J{\'e}gou}]{conneau2017word}
Alexis Conneau, Guillaume Lample, Marc'Aurelio Ranzato, Ludovic Denoyer, and
  Herv{\'e} J{\'e}gou. 2017{\natexlab{b}}.
\newblock Word translation without parallel data.
\newblock \emph{arXiv preprint arXiv:1710.04087}.

\bibitem[{Cui et~al.(2019)Cui, Che, Liu, Qin, Wang, and Hu}]{cui2019cross}
Yiming Cui, Wanxiang Che, Ting Liu, Bing Qin, Shijin Wang, and Guoping Hu.
  2019.
\newblock Cross-lingual machine reading comprehension.
\newblock In \emph{Proceedings of the 2019 Conference on Empirical Methods in
  Natural Language Processing and the 9th International Joint Conference on
  Natural Language Processing (EMNLP-IJCNLP)}, pages 1586--1595.

\bibitem[{Devlin et~al.(2019)Devlin, Chang, Lee, and
  Toutanova}]{devlin2019bert}
Jacob Devlin, Ming-Wei Chang, Kenton Lee, and Kristina Toutanova. 2019.
\newblock Bert: Pre-training of deep bidirectional transformers for language
  understanding.
\newblock In \emph{Proceedings of the 2019 Conference of the North American
  Chapter of the Association for Computational Linguistics: Human Language
  Technologies, Volume 1 (Long and Short Papers)}, pages 4171--4186.

\bibitem[{Dolan and Brockett(2005)}]{dolan2005automatically}
Bill Dolan and Chris Brockett. 2005.
\newblock Automatically constructing a corpus of sentential paraphrases.
\newblock In \emph{Third International Workshop on Paraphrasing (IWP2005)}.

\bibitem[{Fang et~al.(2020)Fang, Wang, Gan, Sun, and Liu}]{fang2020filter}
Yuwei Fang, Shuohang Wang, Zhe Gan, Siqi Sun, and Jingjing Liu. 2020.
\newblock Filter: An enhanced fusion method for cross-lingual language
  understanding.
\newblock \emph{arXiv preprint arXiv:2009.05166}.

\bibitem[{Fluhr et~al.(1999)Fluhr, Frederking, Oard, Okumura, Ishikawa, and
  Satoh}]{fluhr1999multilingual}
Christian Fluhr, Robert~E Frederking, Doug Oard, Akitoshi Okumura, Kai
  Ishikawa, and Kenji Satoh. 1999.
\newblock Multilingual (or cross-lingual) information retrieval.
\newblock \emph{Proceedings of the Multilingual Information Management: Current
  Levels and Future Abilities}.

\bibitem[{Goodfellow et~al.(2014)Goodfellow, Pouget-Abadie, Mirza, Xu,
  Warde-Farley, Ozair, Courville, and Bengio}]{goodfellow2014generative}
Ian Goodfellow, Jean Pouget-Abadie, Mehdi Mirza, Bing Xu, David Warde-Farley,
  Sherjil Ozair, Aaron Courville, and Yoshua Bengio. 2014.
\newblock Generative adversarial nets.
\newblock In \emph{Advances in neural information processing systems}, pages
  2672--2680.

\bibitem[{Group(2011)}]{miniwatts2011}
Miniwatts~Marketing Group. 2011.
\newblock Internet world stats: Usage and population statistics.
\newblock \emph{Miniwatts Marketing Group}.

\bibitem[{Haji\v{c} et~al.(2000)Haji\v{c}, Hric, and
  Kubo\v{n}}]{10.3115/974147.974149}
Jan Haji\v{c}, Jan Hric, and Vladislav Kubo\v{n}. 2000.
\newblock \href {https://doi.org/10.3115/974147.974149} {Machine translation of
  very close languages}.
\newblock In \emph{Proceedings of the Sixth Conference on Applied Natural
  Language Processing}, ANLC '00, page 7–12, USA. Association for
  Computational Linguistics.

\bibitem[{Honnibal and Montani(2017)}]{spacy2}
Matthew Honnibal and Ines Montani. 2017.
\newblock {spaCy 2}: Natural language understanding with {B}loom embeddings,
  convolutional neural networks and incremental parsing.
\newblock To appear.

\bibitem[{Johnson et~al.(2017)Johnson, Schuster, Le, Krikun, Wu, Chen, Thorat,
  Vi{\'e}gas, Wattenberg, Corrado et~al.}]{johnson2017google}
Melvin Johnson, Mike Schuster, Quoc~V Le, Maxim Krikun, Yonghui Wu, Zhifeng
  Chen, Nikhil Thorat, Fernanda Vi{\'e}gas, Martin Wattenberg, Greg Corrado,
  et~al. 2017.
\newblock Google’s multilingual neural machine translation system: Enabling
  zero-shot translation.
\newblock \emph{Transactions of the Association for Computational Linguistics},
  5:339--351.

\bibitem[{Joshi et~al.(2017)Joshi, Choi, Weld, and
  Zettlemoyer}]{joshi2017triviaqa}
Mandar Joshi, Eunsol Choi, Daniel~S Weld, and Luke Zettlemoyer. 2017.
\newblock Triviaqa: A large scale distantly supervised challenge dataset for
  reading comprehension.
\newblock \emph{arXiv preprint arXiv:1705.03551}.

\bibitem[{Karpukhin et~al.(2020)Karpukhin, Oğuz, Min, Lewis, Wu, Edunov, Chen,
  and tau Yih}]{karpukhin2020dense}
Vladimir Karpukhin, Barlas Oğuz, Sewon Min, Patrick Lewis, Ledell Wu, Sergey
  Edunov, Danqi Chen, and Wen tau Yih. 2020.
\newblock \href {http://arxiv.org/abs/2004.04906} {Dense passage retrieval for
  open-domain question answering}.

\bibitem[{Khandelwal et~al.(2020)Khandelwal, Fan, Jurafsky, Zettlemoyer, and
  Lewis}]{khandelwal2020nearest}
Urvashi Khandelwal, Angela Fan, Dan Jurafsky, Luke Zettlemoyer, and Mike Lewis.
  2020.
\newblock Nearest neighbor machine translation.
\newblock \emph{arXiv preprint arXiv:2010.00710}.

\bibitem[{Kwiatkowski et~al.(2019)Kwiatkowski, Palomaki, Redfield, Collins,
  Parikh, Alberti, Epstein, Polosukhin, Devlin, Lee
  et~al.}]{kwiatkowski2019natural}
Tom Kwiatkowski, Jennimaria Palomaki, Olivia Redfield, Michael Collins, Ankur
  Parikh, Chris Alberti, Danielle Epstein, Illia Polosukhin, Jacob Devlin,
  Kenton Lee, et~al. 2019.
\newblock Natural questions: A benchmark for question answering research.
\newblock \emph{Transactions of the Association for Computational Linguistics},
  7:453--466.

\bibitem[{Lample and Conneau(2019)}]{lample2019cross}
Guillaume Lample and Alexis Conneau. 2019.
\newblock Cross-lingual language model pretraining.
\newblock \emph{arXiv preprint arXiv:1901.07291}.

\bibitem[{Lee and Lee(2019)}]{lee2019cross}
Chia-Hsuan Lee and Hung-Yi Lee. 2019.
\newblock Cross-lingual transfer learning for question answering.
\newblock \emph{arXiv preprint arXiv:1907.06042}.

\bibitem[{Lehal(2018)}]{manpreet2018}
Manpreet Lehal. 2018.
\newblock Challenges in cross language information retrieval.

\bibitem[{Lewis et~al.(2019)Lewis, O{\u{g}}uz, Rinott, Riedel, and
  Schwenk}]{lewis2019mlqa}
Patrick Lewis, Barlas O{\u{g}}uz, Ruty Rinott, Sebastian Riedel, and Holger
  Schwenk. 2019.
\newblock Mlqa: Evaluating cross-lingual extractive question answering.
\newblock \emph{arXiv preprint arXiv:1910.07475}.

\bibitem[{Lewis et~al.(2020)Lewis, Stenetorp, and Riedel}]{lewis2020question}
Patrick Lewis, Pontus Stenetorp, and Sebastian Riedel. 2020.
\newblock Question and answer test-train overlap in open-domain question
  answering datasets.
\newblock \emph{arXiv preprint arXiv:2008.02637}.

\bibitem[{Longpre et~al.(2020)Longpre, Lu, and Daiber}]{longpre2020mkqa}
Shayne Longpre, Yi~Lu, and Joachim Daiber. 2020.
\newblock \href {https://arxiv.org/pdf/2007.15207.pdf} {Mkqa: A linguistically
  diverse benchmark for multilingual open domain question answering}.

\bibitem[{Mikolov et~al.(2013)Mikolov, Le, and
  Sutskever}]{mikolov2013exploiting}
Tomas Mikolov, Quoc~V Le, and Ilya Sutskever. 2013.
\newblock Exploiting similarities among languages for machine translation.
\newblock \emph{arXiv preprint arXiv:1309.4168}.

\bibitem[{Mulcaire et~al.(2019)Mulcaire, Kasai, and
  Smith}]{mulcaire2019polyglot}
Phoebe Mulcaire, Jungo Kasai, and Noah~A Smith. 2019.
\newblock Polyglot contextual representations improve crosslingual transfer.
\newblock In \emph{Proceedings of the 2019 Conference of the North American
  Chapter of the Association for Computational Linguistics: Human Language
  Technologies, Volume 1 (Long and Short Papers)}, pages 3912--3918.

\bibitem[{Plank and Agi{\'c}(2018)}]{plank2018distant}
Barbara Plank and {\v{Z}}eljko Agi{\'c}. 2018.
\newblock \href {https://doi.org/10.18653/v1/D18-1061} {Distant supervision
  from disparate sources for low-resource part-of-speech tagging}.
\newblock In \emph{Proceedings of the 2018 Conference on Empirical Methods in
  Natural Language Processing}, pages 614--620, Brussels, Belgium. Association
  for Computational Linguistics.

\bibitem[{Rajpurkar et~al.(2016)Rajpurkar, Zhang, Lopyrev, and
  Liang}]{rajpurkar2016squad}
Pranav Rajpurkar, Jian Zhang, Konstantin Lopyrev, and Percy Liang. 2016.
\newblock Squad: 100,000+ questions for machine comprehension of text.
\newblock In \emph{Proceedings of the 2016 Conference on Empirical Methods in
  Natural Language Processing}, pages 2383--2392.

\bibitem[{Reimers and Gurevych(2019)}]{reimers2019sentence}
Nils Reimers and Iryna Gurevych. 2019.
\newblock Sentence-bert: Sentence embeddings using siamese bert-networks.
\newblock In \emph{Proceedings of the 2019 Conference on Empirical Methods in
  Natural Language Processing and the 9th International Joint Conference on
  Natural Language Processing (EMNLP-IJCNLP)}, pages 3973--3983.

\bibitem[{Reimers and Gurevych(2020)}]{reimers2020making}
Nils Reimers and Iryna Gurevych. 2020.
\newblock Making monolingual sentence embeddings multilingual using knowledge
  distillation.
\newblock \emph{arXiv preprint arXiv:2004.09813}.

\bibitem[{Roy et~al.(2020)Roy, Constant, Al-Rfou, Barua, Phillips, and
  Yang}]{roy2020lareqa}
Uma Roy, Noah Constant, Rami Al-Rfou, Aditya Barua, Aaron Phillips, and Yinfei
  Yang. 2020.
\newblock Lareqa: Language-agnostic answer retrieval from a multilingual pool.
\newblock \emph{arXiv preprint arXiv:2004.05484}.

\bibitem[{Schneider et~al.(2013)Schneider, Mohit, Dyer, Oflazer, and
  Smith}]{schneider-etal-2013-supersense}
Nathan Schneider, Behrang Mohit, Chris Dyer, Kemal Oflazer, and Noah~A. Smith.
  2013.
\newblock \href {https://www.aclweb.org/anthology/N13-1076} {Supersense tagging
  for {A}rabic: the {MT}-in-the-middle attack}.
\newblock In \emph{Proceedings of the 2013 Conference of the North {A}merican
  Chapter of the Association for Computational Linguistics: Human Language
  Technologies}, pages 661--667, Atlanta, Georgia. Association for
  Computational Linguistics.

\bibitem[{Schuster et~al.(2019)Schuster, Ram, Barzilay, and
  Globerson}]{schuster2019cross}
Tal Schuster, Ori Ram, Regina Barzilay, and Amir Globerson. 2019.
\newblock Cross-lingual alignment of contextual word embeddings, with
  applications to zero-shot dependency parsing.
\newblock In \emph{Proceedings of the 2019 Conference of the North American
  Chapter of the Association for Computational Linguistics: Human Language
  Technologies, Volume 1 (Long and Short Papers)}, pages 1599--1613.

\bibitem[{Sharma et~al.(2019)Sharma, Graesser, Nangia, and
  Evci}]{sharma2019natural}
Lakshay Sharma, Laura Graesser, Nikita Nangia, and Utku Evci. 2019.
\newblock Natural language understanding with the quora question pairs dataset.
\newblock \emph{arXiv preprint arXiv:1907.01041}.

\bibitem[{Vaswani et~al.(2017)Vaswani, Shazeer, Parmar, Uszkoreit, Jones,
  Gomez, Kaiser, and Polosukhin}]{vaswani2017attention}
Ashish Vaswani, Noam Shazeer, Niki Parmar, Jakob Uszkoreit, Llion Jones,
  Aidan~N Gomez, {\L}ukasz Kaiser, and Illia Polosukhin. 2017.
\newblock Attention is all you need.
\newblock In \emph{Proceedings of the 31st International Conference on Neural
  Information Processing Systems}, pages 6000--6010.

\bibitem[{Vrande\v{c}i\'{c} and Kr\"{o}tzsch(2014)}]{10.1145/2629489}
Denny Vrande\v{c}i\'{c} and Markus Kr\"{o}tzsch. 2014.
\newblock \href {https://doi.org/10.1145/2629489} {Wikidata: A free
  collaborative knowledgebase}.
\newblock \emph{Commun. ACM}, 57(10):78–85.

\bibitem[{Wang et~al.(2017)Wang, Hamza, and Florian}]{wang2017bilateral}
Zhiguo Wang, Wael Hamza, and Radu Florian. 2017.
\newblock Bilateral multi-perspective matching for natural language sentences.
\newblock In \emph{Proceedings of the 26th International Joint Conference on
  Artificial Intelligence}, pages 4144--4150.

\bibitem[{Wolf et~al.(2019)Wolf, Debut, Sanh, Chaumond, Delangue, Moi, Cistac,
  Rault, Louf, Funtowicz, and Brew}]{Wolf2019HuggingFacesTS}
Thomas Wolf, Lysandre Debut, Victor Sanh, Julien Chaumond, Clement Delangue,
  Anthony Moi, Pierric Cistac, Tim Rault, R'emi Louf, Morgan Funtowicz, and
  Jamie Brew. 2019.
\newblock Huggingface's transformers: State-of-the-art natural language
  processing.
\newblock \emph{ArXiv}, abs/1910.03771.

\bibitem[{Wu and Dredze(2020)}]{wu-dredze-2020-languages}
Shijie Wu and Mark Dredze. 2020.
\newblock \href {https://doi.org/10.18653/v1/2020.repl4nlp-1.16} {Are all
  languages created equal in multilingual {BERT}?}
\newblock In \emph{Proceedings of the 5th Workshop on Representation Learning
  for NLP}, pages 120--130, Online. Association for Computational Linguistics.

\bibitem[{Xie et~al.(2018)Xie, Yang, Neubig, Smith, and
  Carbonell}]{xie2018neural}
Jiateng Xie, Zhilin Yang, Graham Neubig, Noah~A. Smith, and Jaime Carbonell.
  2018.
\newblock \href {https://doi.org/10.18653/v1/D18-1034} {Neural cross-lingual
  named entity recognition with minimal resources}.
\newblock In \emph{Proceedings of the 2018 Conference on Empirical Methods in
  Natural Language Processing}, pages 369--379, Brussels, Belgium. Association
  for Computational Linguistics.

\bibitem[{Yang et~al.(2019{\natexlab{a}})Yang, Cer, Ahmad, Guo, Law, Constant,
  Abrego, Yuan, Tar, Sung et~al.}]{yang2019multilingual}
Yinfei Yang, Daniel Cer, Amin Ahmad, Mandy Guo, Jax Law, Noah Constant,
  Gustavo~Hernandez Abrego, Steve Yuan, Chris Tar, Yun-Hsuan Sung, et~al.
  2019{\natexlab{a}}.
\newblock Multilingual universal sentence encoder for semantic retrieval.
\newblock \emph{arXiv preprint arXiv:1907.04307}.

\bibitem[{Yang et~al.(2019{\natexlab{b}})Yang, Zhang, Tar, and
  Baldridge}]{yang2019paws}
Yinfei Yang, Yuan Zhang, Chris Tar, and Jason Baldridge. 2019{\natexlab{b}}.
\newblock Paws-x: A cross-lingual adversarial dataset for paraphrase
  identification.
\newblock In \emph{Proceedings of the 2019 Conference on Empirical Methods in
  Natural Language Processing and the 9th International Joint Conference on
  Natural Language Processing (EMNLP-IJCNLP)}, pages 3678--3683.

\bibitem[{Zhang et~al.(2020)Zhang, Williams, Titov, and
  Sennrich}]{zhang2020improving}
Biao Zhang, Philip Williams, Ivan Titov, and Rico Sennrich. 2020.
\newblock Improving massively multilingual neural machine translation and
  zero-shot translation.
\newblock \emph{arXiv preprint arXiv:2004.11867}.

\bibitem[{Zhang et~al.(2019)Zhang, Baldridge, and He}]{zhang2019paws}
Yuan Zhang, Jason Baldridge, and Luheng He. 2019.
\newblock Paws: Paraphrase adversaries from word scrambling.
\newblock In \emph{Proceedings of the 2019 Conference of the North American
  Chapter of the Association for Computational Linguistics: Human Language
  Technologies, Volume 1 (Long and Short Papers)}, pages 1298--1308.

\bibitem[{Zitouni and Florian(2008)}]{zitouni-florian-2008-mention}
Imed Zitouni and Radu Florian. 2008.
\newblock \href {https://www.aclweb.org/anthology/D08-1063} {Mention detection
  crossing the language barrier}.
\newblock In \emph{Proceedings of the 2008 Conference on Empirical Methods in
  Natural Language Processing}, pages 600--609, Honolulu, Hawaii. Association
  for Computational Linguistics.

\end{thebibliography}
\bibliographystyle{acl_natbib}

\clearpage
\appendix

\label{sec:appendix}
\setlength{\abovecaptionskip}{6pt}
\setlength{\belowcaptionskip}{-6pt}


\section{Reproducibility}




\subsection{Experimental Setup}

\paragraph{Computing Infrastructure.}
For all of our experiments, we used a computation cluster with 4 NVIDIA Tesla V100 GPUs, 32GB GPU memory and 256GB RAM.

\paragraph{Implementation}
We used Python 3.7, PyTorch 1.4.0, and Transformers 2.8.0 for all our experiments. We obtain our datasets from the citations specified in the main paper, and link to the repositories of all libraries we use.


\paragraph{Hyperparameter Search}
For our hyper parameter searches, we perform a uniformly random search over learning rate and batch size, with ranges specified in Table ~\ref{tab:xxencomp}, optimizing for the development accuracy. We find the optimal learning rate and batch size pair to be $1e-5$ and $80$ respectively.

\paragraph{Evaluation}
For query matching, we use scikit-learn \footnote{\url{https://scikit-learn.org/stable/}} to calculate the accuracy. For end-to-end performance, we use the MLQA evaluation script to obtain the F1 score of the results\footnote{\url{https://github.com/facebookresearch/MLQA}}.











\paragraph{Datasets}
We use the sentences in each dataset as-is, and rely on the pretrained tokenizer for each model to perform preprocessing.


\subsection{Model Training}

\paragraph{Query Paraphrase Dataset}
We found the optimal training combination of the PAWS-X and QQP datasets by training XLM-R classifiers on training dataset percentages of $(100\%, 0\%)$, $(75\%, 25\%)$, and $(50\%, 50\%)$ of (PAWS-X, QQP) -- with the PAWS-X percentage entailing the entirety of the PAWS-X dataset -- and observe the performance on matching multilingual XQuAD queries. 
We shuffle the examples in the training set, and restrict the input examples to being (English, LRL) pairs. 
We perform a hyperparameter search as specified in Table ~\ref{tab:ceparams} for each dataset composition, and report the test results in Table ~\ref{tab:xxencomp}.

\begin{table}[h]
    \centering
    \begin{tabular}{c|c}
        (PAWS-X, QQP) & XQuAD \\\hline
        $(100\%, 0\%)$ & 0.847  \\
        $(75\%, 25\%)$ & \textbf{0.985}\\
        $(50\%, 50\%)$ & 0.979
    \end{tabular}
    \caption{\label{tab:xxencomp}\textbf{XLM-R Query Paraphrase Performance On Different Query Compositions.} The performance of XLM-Roberta on matching XQuAD test queries when finetuned on different training set compositions of PAWS-X and QQP.}
\end{table}

\subsection{Cross Encoder}
We start with the pretrained \texttt{xlm-roberta-large} checkpoint in Huggingface's transformers\footnote{\url{https://github.com/huggingface/transformers}} library and perform a hyperparameter search with the parameters specified in Table 1 by using a modified version of Huggingface's text classification training pipeline for GLUE.

The cross encoder was used in all the RM-MIPS methods. In particular, it was used in the RM-MIPS (mUSE), RM-MIPS (LASER), and RM-MIPS (XLM-R) rows of tables in the main paper.

\begin{table}[h]
\small
\centering
\begin{tabular}{ll}
\toprule
\textsc{Model Parameters} & \textsc{Value/Range} \\
\midrule
\textbf{Fixed Parameters} & {} \\
\midrule
Model & XLM-Roberta Large \\
Num Epochs & 3 \\
Dropout & 0.1 \\
Optimizer & Adam \\
Learning Rate Schedule & Linear Decay \\
Max Sequence Length & 128 \\
\midrule
\textbf{Tuned Parameters} & {} \\
\midrule
Batch Size & [8, 120] \\
Learning Rate & [$9e-4$, $1e-6$] \\
\midrule
\textbf{Extra Info} & {} \\
\midrule
Model Size (\# params) & $550M$ \\
Vocab Size & 250,002 \\
Trials & 30 \\
\bottomrule
\end{tabular}
\caption{\label{tab:ceparams}
\textbf{Cross Encoder Hyperparameter Selection And Tuning Ranges} The hyper parameters we chose and searched over for XLM-Roberta large on the query paraphrase detection datasets.
}
\end{table}


\section{Full Results Breakdowns}




\subsection{LRL$\rightarrow$HRL Results}
See Table ~\ref{tab:mkqatoen} and ~\ref{tab:xquadtoen} for the non-aggregated LRL$\rightarrow$HRL language performances of each method on MKQA and XQuAD respectively.


\subsection{LRL$\rightarrow$HRL$\rightarrow$LRL Results}
See Table ~\ref{tab:mkqaendtoend} and ~\ref{tab:xquadendtoend} for the non-aggregated LRL$\rightarrow$HRL$\rightarrow$LRL language performances of each method on MKQA and XQuAD respectively.


\begin{table*}[htbp]
\centerline{
\begin{small}
\begin{tabular}{l|c|c|c|c|c|c|c|c|c|c|c|c|c}
\toprule
 & ar & $\text{zh}_\text{cn}$ & da & de & es & fi & fr & he & $\text{zh}_\text{hk}$ & hu & it & ja & km \\
\midrule
NMT + MIPS & 69.2 & 48.0 & 89.8 & 87.5 & 86.5 & 76.0 & 87.6 & 74.3 & 42.5 & 79.1 & 86.6 & 62.0 & 45.4 \\
mUSE & 80.0 & 83.2 & 51.7 & 90.9 & 91.7 & 37.6 & 91.5 & 33.5 & 80.8 & 40.7 & 91.6 & 80.0 & 35.6 \\
LASER & 81.5 & 62.8 & 88.6 & 52.0 & 79.9 & 81.6 & 78.5 & 85.5 & 64.0 & 69.1 & 80.4 & 39.3 & 40.2 \\
Single Encoder (XLM-R) & 58.0 & 76.3 & 84.8 & 74.6 & 73.3 & 65.5 & 74.1 & 67.8 & 77.0 & 66.9 & 69.0 & 71.4 & 59.0 \\
RM-MIPS (mUSE) & 77.6 & 81.2 & 77.2 & 88.8 & 88.9 & 59.9 & 88.8 & 44.1 & 81.2 & 64.1 & 88.4 & 81.2 & 50.6 \\
RM-MIPS (LASER) & 77.2 & 77.7 & 89.2 & 66.9 & 84.7 & 84.8 & 84.4 & 83.3 & 78.1 & 77.5 & 84.7 & 64.2 & 48.2 \\
RM-MIPS (Ours) & 72.6 & 80.7 & 90.1 & 86.8 & 87.0 & 82.7 & 87.2 & 80.6 & 81.0 & 81.4 & 85.5 & 79.5 & 72.4 \\
\midrule \midrule
 & ko & ms & nl & no & pl & pt & ru & sv & th & tr & $\text{zh}_\text{tw}$ & vi \\
 \midrule
NMT + MIPS  & 54.2 & 86.0 & 88.8 & 87.2 & 81.9 & 87.4 & 81.9 & 87.2 & 75.0 & 79.6 & 39.7 & 76.0 \\
mUSE  & 73.7 & 87.6 & 92.0 & 50.3 & 84.9 & 93.3 & 87.2 & 50.3 & 88.6 & 87.0 & 73.2 & 38.6 \\
LASER  & 68.6 & 92.5 & 93.1 & 92.4 & 73.7 & 85.2 & 78.1 & 92.8 & 62.1 & 75.2 & 57.9 & 79.9 \\
Single Encoder (XLM-R)  & 72.3 & 76.4 & 79.1 & 81.3 & 70.6 & 65.7 & 78.8 & 83.8 & 79.4 & 68.7 & 71.2 & 78.6 \\
RM-MIPS (mUSE)  & 74.7 & 89.9 & 90.9 & 75.6 & 87.3 & 89.8 & 87.1 & 76.0 & 86.8 & 86.6 & 75.0 & 64.4 \\
RM-MIPS (LASER)  & 73.1 & 89.5 & 90.2 & 89.7 & 81.9 & 86.7 & 84.3 & 89.8 & 77.0 & 82.3 & 72.5 & 84.0 \\
RM-MIPS (XLM-R)  & 75.2 & 89.0 & 89.8 & 88.8 & 85.6 & 85.5 & 86.1 & 90.0 & 85.4 & 83.6 & 75.5 & 85.9 \\

\bottomrule
\end{tabular}
\end{small}}
\caption{\label{tab:mkqatoen}\textbf{MKQA + Natural Questions Per-Language LRL$\rightarrow$HRL Results.} The accuracy scores for each method on query matching.}
\end{table*}

\begin{table*}[htbp]
\centering
\begin{small}
\begin{tabular}{l|c|c|c|c|c|c|c|c|c|c}
\toprule
 & ar & de & el & es & hi & ru & th & tr & vi & zh \\
\midrule
NMT + MIPS & 71.7 & 90.8 & 86.7 & 95.2 & 79.9 & 85.7 & 67.4 & 82.9 & 74.8 & 41.8 \\
mUSE & 87.4 & 96.4 & 7.5 & 98.1 & 3.4 & 93.2 & 91.6 & 94.1 & 17.8 & 90.3 \\
LASER & 61.7 & 33.1 & 3.7 & 86.2 & 28.6 & 70.4 & 24.2 & 65.3 & 64.7 & 29.2 \\
Single Encoder (XLM-R) & 66.8 & 85.1 & 81.7 & 87.8 & 77.6 & 85.0 & 76.6 & 81.9 & 89.4 & 82.3 \\
RM-MIPS (mUSE) & 90.4 & 96.3 & 14.8 & 97.3 & 10.1 & 93.2 & 92.6 & 95.7 & 39.3 & 91.0 \\
RM-MIPS (LASER) & 81.6 & 59.9 & 11.1 & 95.5 & 59.1 & 88.3 & 55.5 & 89.0 & 85.7 & 66.2 \\
RM-MIPS (XLM-R) & 86.6 & 94.2 & 94.1 & 95.5 & 92.0 & 93.0 & 90.7 & 92.5 & 92.1 & 90.8 \\
\bottomrule
\end{tabular}
\end{small}
\caption{\label{tab:xquadtoen}\textbf{XQuAD + SQuAD Per-Language LRL$\rightarrow$HRL Results.} The accuracy scores for each method on query matching.}
\end{table*}

\newpage

\begin{table*}[htbp]
\centerline{
\begin{small}
\begin{tabular}{l|c|c|c|c|c|c|c|c|c|c|c|c|c}
\toprule
 & ar & $\text{zh}_\text{cn}$ & da & de & es & fi & fr & he & $\text{zh}_\text{hk}$ & hu & it & ja & km  \\
\midrule
NMT + MIPS & 60.0 & 41.7 & 85.8 & 83.8 & 82.4 & 72.0 & 83.7 & 63.3 & 41.2 & 74.5 & 82.5 & 60.1 & 44.8 \\
mUSE & 68.6 & 62.7 & 50.1 & 87.2 & 87.4 & 37.2 & 87.5 & 31.9 & 68.7 & 40.0 & 87.2 & 74.9 & 35.0 \\
LASER & 70.1 & 49.5 & 84.6 & 50.8 & 76.3 & 77.3 & 75.3 & 72.8 & 56.2 & 65.0 & 76.8 & 39.1 & 38.1 \\
Single Encoder (XLM-R) & 50.9 & 57.5 & 81.0 & 71.7 & 70.2 & 62.0 & 70.9 & 58.6 & 65.8 & 63.1 & 66.0 & 68.0 & 54.9 \\
RM-MIPS (mUSE) & 66.9 & 61.3 & 74.4 & 85.2 & 84.8 & 58.0 & 84.9 & 39.9 & 68.8 & 61.5 & 84.1 & 75.8 & 46.0 \\
RM-MIPS (LASER) & 66.7 & 59.0 & 85.0 & 64.6 & 80.7 & 80.3 & 80.6 & 71.0 & 66.3 & 72.7 & 80.6 & 61.7 & 45.3 \\
RM-MIPS (Ours) & 62.8 & 60.8 & 85.9 & 83.3 & 83.1 & 78.4 & 83.3 & 68.7 & 68.6 & 76.6 & 81.5 & 74.4 & 64.4 \\
\midrule\midrule
& ko & ms & nl & no & pl & pt & ru & sv & th & tr & $\text{zh}_\text{tw}$ & vi &  \\
\midrule
NMT + MIPS  & 47.5 & 81.1 & 85.3 & 80.2 & 77.6 & 83.3 & 72.6 & 84.1 & 62.9 & 74.7 & 35.2 & 70.6 & \\
mUSE  & 63.0 & 82.7 & 88.5 & 48.4 & 80.4 & 88.9 & 77.2 & 49.4 & 72.2 & 81.7 & 55.6 & 37.7 & \\
LASER  & 59.1 & 87.4 & 89.7 & 85.1 & 70.0 & 81.2 & 69.4 & 89.5 & 53.7 & 70.7 & 45.7 & 74.4 & \\
Single Encoder (XLM-R)  & 62.5 & 72.2 & 76.0 & 75.2 & 67.0 & 62.5 & 70.1 & 80.8 & 66.3 & 64.6 & 54.1 & 73.1 & \\
RM-MIPS (mUSE)  & 64.2 & 84.8 & 87.3 & 70.6 & 82.5 & 85.3 & 77.1 & 73.9 & 70.7 & 81.2 & 56.6 & 61.3 & \\
RM-MIPS (LASER)  & 63.1 & 84.4 & 86.7 & 81.8 & 77.3 & 82.4 & 74.7 & 86.6 & 64.3 & 77.1 & 55.1 & 78.2 & \\
RM-MIPS (XLM-R)  & 64.7 & 84.0 & 86.3 & 81.6 & 81.0 & 81.3 & 76.3 & 86.9 & 69.9 & 78.6 & 56.8 & 79.9 & \\
\bottomrule
\end{tabular}
\end{small}}
\caption{\label{tab:mkqaendtoend}\textbf{MKQA + Natural Questions Per-Language LRL$\rightarrow$HRL$\rightarrow$LRL WikiData Results.} The F1 scores for end-to-end performance of each method on every language when using WikiData translation}
\end{table*}

\begin{table*}[htbp]
\centering
\begin{small}
\begin{tabular}{l|c|c|c|c|c|c|c|c|c|c}
\toprule
 & ar & de & el & es & hi & ru & th & tr & vi & zh \\
\midrule
NMT + MIPS & 35.3 & 55.5 & 39.2 & 68.2 & 32.9 & 30.7 & 17.8 & 42.1 & 45.6 & 19.0 \\
mUSE & 40.8 & 58.2 & 4.4 & 70.0 & 1.6 & 33.4 & 23.4 & 47.0 & 11.8 & 33.6 \\
LASER & 29.9 & 22.7 & 1.5 & 61.8 & 10.8 & 24.2 & 6.4 & 33.0 & 38.6 & 12.7 \\
Single Encoder (XLM-R) & 31.3 & 52.9 & 37.3 & 63.9 & 30.9 & 30.1 & 18.6 & 42.0 & 52.7 & 30.6 \\
RM-MIPS (mUSE) & 42.6 & 58.1 & 7.8 & 69.6 & 4.2 & 33.4 & 23.2 & 47.5 & 26.1 & 33.8 \\
RM-MIPS (LASER) & 38.3 & 38.2 & 5.7 & 68.3 & 22.9 & 31.1 & 13.7 & 44.5 & 50.7 & 26.3 \\
RM-MIPS (XLM-R) & 40.9 & 57.3 & 42.1 & 68.7 & 36.7 & 33.0 & 22.9 & 45.7 & 54.5 & 33.6 \\
\bottomrule
\end{tabular}
\end{small}
\caption{\label{tab:xquadendtoend}\textbf{XQuAD + SQuAD Per-Language LRL$\rightarrow$HRL$\rightarrow$LRL NMT Results.} The F1 scores for end-to-end performance of each method on every language when using NMT translation}
\end{table*}

\end{document}